\documentclass[sigconf,screen,nonacm,language=english]{acmart}

\usepackage{algorithm}
\usepackage{algpseudocode}
\usepackage{subcaption}
\usepackage{multirow}

\AtBeginDocument{%
  \providecommand\BibTeX{{%
    \normalfont B\kern-0.5em{\scshape i\kern-0.25em b}\kern-0.8em\TeX}}}

\setcopyright{acmlicensed}
\copyrightyear{2024}
\acmYear{2024}
\acmDOI{}

\acmConference[]{}{}{}
\acmISBN{}

\begin{document}

\title{Continuous fake media detection: adapting deepfake detectors to new generative techniques}

\author[1]{Francesco Tassone}
\email{tassone.1814263@studenti.uniroma1.it}
\affiliation{
 \institution{Sapienza University of Rome}
 \streetaddress{Via Ariosto 25}
 \city{Rome}
 \state{Rome}
 \country{Italy}
 \postcode{00185}
}

\author[1,2]{Luca Maiano}
\email{maiano@diag.uniroma1.it}
\affiliation{
 \institution{Sapienza University of Rome}
 \streetaddress{Via Ariosto 25}
 \city{Rome}
 \state{Rome}
 \country{Italy}
 \postcode{00185}
}
\affiliation{
 \institution{Ubiquitous srl}
 \streetaddress{Via Ariosto 25}
 \city{Rome}
 \state{Rome}
 \country{Italy}
 \postcode{00185}
}

\author{Irene Amerini}
\email{amerini@diag.uniroma1.it}
\affiliation{
 \institution{Sapienza University of Rome}
 \streetaddress{Via Ariosto 25}
 \city{Rome}
 \state{Rome}
 \country{Italy}
 \postcode{00185}
}

\renewcommand{\shortauthors}{Tassone et al.}

\begin{abstract}
    Generative techniques continue to evolve at an impressively high rate, driven by the hype about these technologies. This rapid advancement severely limits the application of deepfake detectors, which, despite numerous efforts by the scientific community, struggle to achieve sufficiently robust performance against the ever-changing content. To address these limitations, in this paper, we propose an analysis of two continuous learning techniques on a \emph{Short} and a \emph{Long} sequence of fake media. Both sequences include a complex and heterogeneous range of deepfakes (generated images and videos) from GANs, computer graphics techniques, and unknown sources. 

    Our experiments show that continual learning could be important in mitigating the need for generalizability. In fact, we show that, although with some limitations, continual learning methods help to maintain good performance across the entire training sequence. For these techniques to work in a sufficiently robust way, however, it is necessary that the tasks in the sequence share similarities. In fact, according to our experiments, the order and similarity of the tasks can affect the performance of the models over time. To address this problem, we show that it is possible to group tasks based on their similarity. This small measure allows for a significant improvement even in longer sequences. This result suggests that continual techniques can be combined with the most promising detection methods, allowing them to catch up with the latest generative techniques. 
    
    In addition to this, we propose an overview of how this learning approach can be integrated into a deepfake detection pipeline for continuous integration and continuous deployment (CI/CD). This allows you to keep track of different funds, such as social networks, new generative tools, or third-party datasets, and through the integration of continuous learning, allows constant maintenance of the detectors.
\end{abstract}



\maketitle

\section{Introduction}

Generative AI tools like Midjourney\footnote{\url{https://www.midjourney.com/}}, ChatGPT\footnote{\url{https://openai.com/chatgpt}}, or the more recent Sora\footnote{\url{https://openai.com/sora}} are completely revolutionizing the way media content is created, leading to the mass adoption of tools that were unimaginable until recently. However, this progress makes the threat of new-generation disinformation or defamation campaigns increasingly concrete. Unfortunately, forensic tools for detecting this content are not advancing at the same rate. Although it is possible to train highly accurate detectors, these methodologies still poorly generalize to new generative methods due to data drift~\cite{Paleyes2022}. Detectors perform well on the generative techniques they are trained on but commonly fail when exposed to content generated with a new generative model.

Because of these limitations, the practical application of automatic detectors has been almost nil. When someone wants to deploy these tools in commercial or mass verification systems, they must face numerous challenges that go far beyond the need to generalize from a few known benchmarks~\cite{Rossler20191,corvi2023intriguing,Li20231339,Zi20202382}. Prominent among these is the need to continuously train these models on new generative techniques through continuous learning methods.
To address this continuous change, we analyze continuous learning techniques to evaluate their limitations when applied to mitigate this problem. Continual learning, also known as lifelong learning or incremental learning, is a constant learning approach. Unlike transfer learning techniques, where a model trained on one task is retrained on a new task to improve its performance on the latter, continual learning involves maintaining good model performance on a set of evolving tasks as these become available without incurring in \emph{catastrophic forgetting}~\cite{French1999128}. These requirements fit well with the problem of learning to recognize content generated by new techniques and the need to continually readjust a model with respect to data drift produced by the shift in the distribution of data observed in inference versus those seen in training.

This work aims to investigate the effectiveness of two continuous learning techniques with the intention of integrating them into a real, end-to-end deepfake detection system that allows for continuous integration and continuous delivery/deployment (CI/CD). Our goal is therefore to propose a simple yet effective design for a Machine Learning Model Operations (\emph{MLOps}~\cite{Paleyes2022,semola2022continual}) pipeline that would enable the end-to-end development of continuously trained and monitored intelligent detectors with a minimal set of components.

\textbf{Contributions}
Therefore, the main contributions of this preliminary study are summarized below.
\begin{itemize}
    \item \emph{Analysis of continual learning methods.} We study the effectiveness of two continuous learning methods, \emph{Knowledge Distillation}~\cite{hinton2015distilling} (KD) and \emph{Elastic Weight Consolidation}~\cite{Kirkpatrick20173521} (EWC), and show their superiority to transfer learning when continuous training is needed. 
    \item \emph{Sequence.} We study how the order of arrival of the tasks can affect the performance of the model. In particular, we show how task similarity plays an important role in maintaining optimal performance.
    \item \emph{Multi-task continual training.} We show how aggregating tasks based on their similarity can significantly improve the overall performance over the entire sequence. This result is particularly important and helps us to better outline the possible developments of these techniques for deepfake recognition.
    \item \emph{CI/CD pipeline for deepfake detection.} We propose an overview of an end-to-end system for continuous integration and continuous delivery/deployment for a deepfake detection application. 
\end{itemize}

The rest of this paper is organized as follows. Section~\ref{sec:related} offers an overview of the state of the art. In Section~\ref{sec:method}, we introduce the methodology used in this study. In Section~\ref{sec:experiment}, we present the experiments we conducted. Finally, in Section~\ref{sec:conclusion}, we draw the final considerations and illustrate the future developments of this work.

\section{Related works}
\label{sec:related}

Despite the difficulties in keeping up with the advancement of generative techniques, numerous studies have been proposed on detecting generated images and videos~\cite{Amerini2021309,Verdoliva2020910}. In this section, we provide an overview of the most recent advances. However, it is essential to note that many can be combined with the continuous learning techniques analyzed in this paper. In fact, this type of learning approach could be used not as an alternative to these methods but to make these techniques maintainable over time.

First of all, several studies have found that there are some key ingredients for more robust detection~\cite{wang2020cnn,gragnaniello2021gan}. Among them, image compression and resizing can severely mitigate model performance~\cite{papa2023use}. Therefore, to cope with these problems, it is usually recommended to avoid resizes, as they entail image resampling and interpolation, which may erase the subtle high-frequency traces left by the generation process and train models with different forms of augmentation. Moreover, working on local patches also appears to be important~\cite{chai2020makes} as well as analyzing both local and global features~\cite{ju2022fusing}.

A recent study from Aghasanli et al.~\cite{aghasanli2023interpretable} showed that foundational models like ViT can effectively distinguish between authentic and counterfeit images, even when interpretability through prototypes is important. Additionally, the study demonstrated that classifiers with fine-tuned features consistently outperform those utilizing pre-trained weights when applied to cross-dataset domains. Another study by Le et al.~\cite{le2023quality} considers quality factors to train robust detectors. The authors used an intra-model collaborative learning method to minimize the geometrical differences of images in various qualities at different intermediate layers. This idea, combined with an adversarial weight perturbation module, can be used to improve the robustness of the model against input image compression.
Many recent studies also focus on combining different modes, such as audio and video~\cite{raza2023multimodaltrace,zhang2024joint} or video and depth~\cite{maiano2022depthfake}, as well as open-set recognition~\cite{wang2023open}.
 
Other studies focus on reconstructing fake artifacts introduced by generative models by considering second-order statistics in the spatial and frequency domains~\cite{wang2020cnn}. Corvi et al.~\cite{corvi2023intriguing} showed that, similar to GANs, diffusion models also give rise to visible artifacts in the Fourier domain and exhibit anomalous regular patterns in
autocorrelation. In fact, synthetic and real images exhibit significant differences in the mid-high frequency signal content, observable in their radial and angular spectral power distributions.

Among the various detection strategies, watermarks have also been proposed~\cite{fei2023robust,wu2020watermarking}. Through the addition of special information within the image being generated, these watermarks can be used to verify the generative model used to create content. For example, Zhao et al.~\cite{zhao2023proactive} proposed an encoder-decoder network to embed watermarks as anti-deepfake labels into the facial identity features. The injected label is entangled with the facial identity feature, so it will be sensitive to face swap translations and robust to conventional image modifications like resizing and compression. However, these solutions are limited in that they require a model to integrate the watermark into the content during the generation phase.
 
Our method, instead, falls into a different line of studies targeted at designing continual learning methods for deepfake detection. This learning approach has only been partially explored for this task. The study by Marra et al.~\cite{marra2019incremental} was one of the first to propose a multi-task incremental learning method for GAN-generated images based on iCaRL~\cite{rebuffi2017icarl}. A similar approach was applied to videos in Khan et al.~\cite{khan2021video}, while
Pan et al.~\cite{Pan20238035} have recently proposed to learn semantically consistent representations across domains based on supervised contrastive learning and a carefully designed replay set. In contrast, Kim et al.~\cite{kim2021cored} combined a knowledge distillation method with a representation learning loss. 
However, these methods have been tested on datasets not explicitly designed for this type of learning approach, which very often consists of a limited number of generative techniques. To overcome this limitation, Li et al.~\cite{Li20231339} recently introduced a new collection of deepfakes from known and unknown generative models. The proposed CDDB dataset includes multiple evaluations on detecting an easy, hard, and long sequence of deepfakes with appropriate measures.
For these reasons, in this paper, we focus our studies on this dataset by measuring the performance of the continual learning methods examined in this paper on this dataset. Different from other studies, we analyze this specific learning method to integrate it into a CI/CD system appropriately designed for this task in the future.

\begin{figure*}
    \centering
    \includegraphics[width=0.9\linewidth]{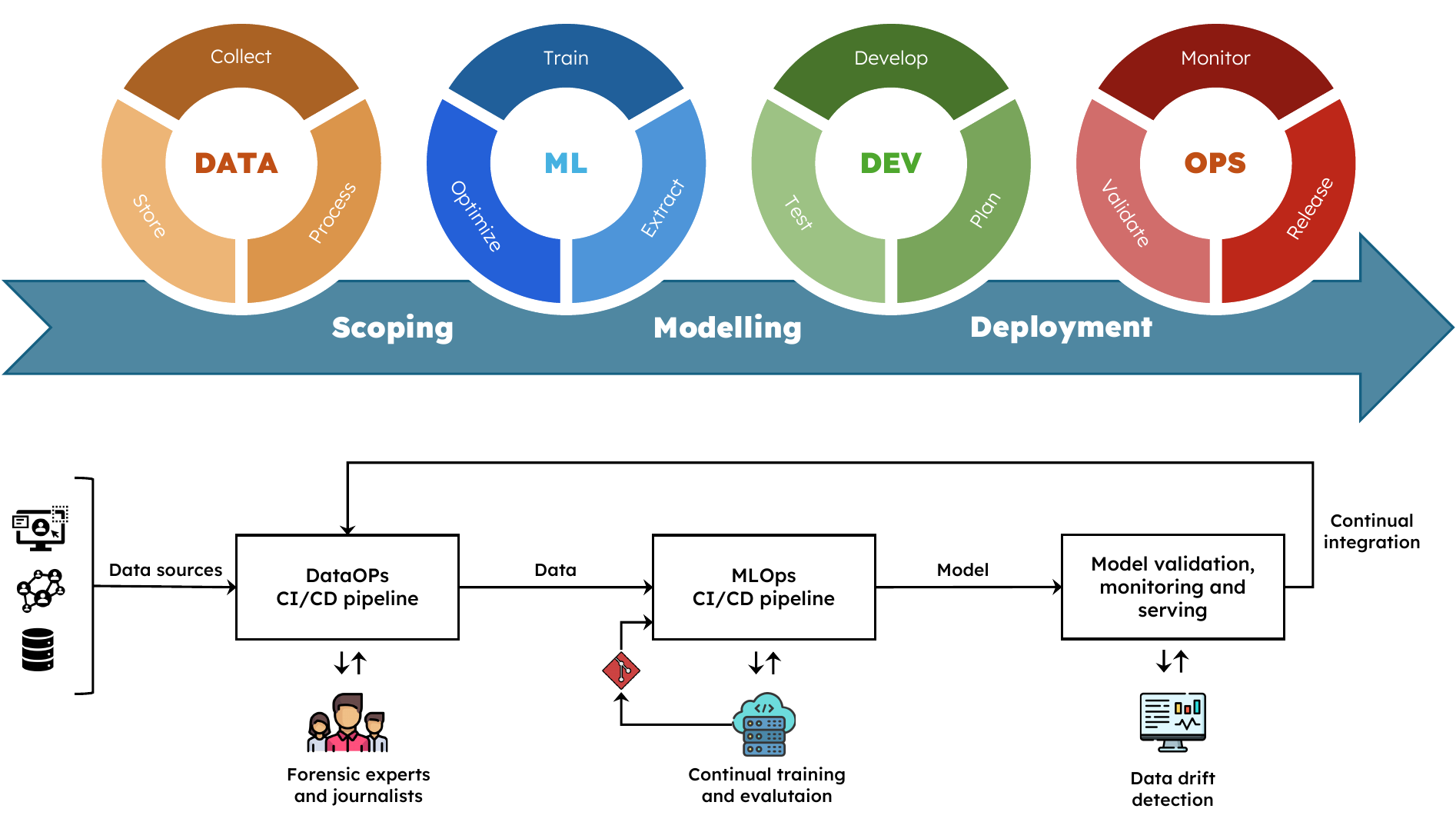}
    \caption{Proposed CI/CD pipeline for deepfake detection. The data coming from different sources like new generative tools, social media or existing databases are analyzed by forensic experts for continual retraining of the system. Next, these data are used for continual learning and monitoring. The data drift distribution module rise an alert whenever it detects new input data distributions. The continaul learning methods analyzed in this paper are part of the \emph{MLOps CI/CD pipeline} block in the figure.}
    \label{fig:mlops_pipeline}
\end{figure*}

\section{Method}
\label{sec:method}

The practical application of deepfake detectors has been severely limited by the need to develop robust detectors with respect to new generative techniques. To overcome this problem, we propose an analysis of two continual learning strategies and an overview of a simple CI/CD pipeline that can complement this preliminary study. This pipeline (depicted in Figure~\ref{fig:mlops_pipeline}) has the advantage of being constantly updated with respect to the latest detectors, which can be adapted for continuous learning as described below.

\subsection{Learning strategies}
\label{sec:learning_strategies}
To evaluate the effectiveness of continual learning in the deepfake recognition task we examine two learning methods. These techniques were chosen based on their demonstrated effectiveness in other tasks~\cite{DeLange20223366}.

\subsubsection*{\textbf{Knowledge Distillation (KD)}}
\label{sec:kd}
Distillation techniques were introduced by Hinton et al.~\cite{hinton2015distilling} in order to transfer knowledge from a neural network $\mathcal{T}$ (the \emph{teacher}) to a neural network $\mathcal{S}$ (the \emph{student}). The key idea behind knowledge distillation is that soft probabilities predicted by a network of trained "teachers" contain much more information about a data point than a simple class label. For example, if multiple classes are assigned high probabilities for an image, this could mean that the image must be close to a decision boundary between those classes. Forcing a student to mimic these probabilities should then cause the student network to absorb some of this knowledge that the teacher discovered above and beyond the information in training labels alone. To implement the KD strategy, we modify the cross-entropy loss $\mathcal{L}_{S}$ by adding a regularization term ($\mathcal{L}_{D}$) as follows.
\begin{equation}
    \mathcal{L}_{KD}(\theta)=
    \alpha \mathcal{L}_{D} + \beta \mathcal{L}_{S}
    \label{eq:kd_loss}
\end{equation}
Here, $\alpha$ and $\beta$ are scalar coefficients that control the balancing between the current and past tasks. The distillation loss $\mathcal{L}_{D}$ uses the output of the teacher model to facilitate knowledge transfer from previous tasks to the new student model. As the teacher is non-trainable, its predictions are solely based on prior knowledge, producing soft labels for the training process of the student model. This term is computed as:
\begin{equation}
    \mathcal{L}_{D}=
    \sum_{x_i \in X}
    \sigma 
    \left( 
    \mathcal{T}(x_i,\hat{y}_i), \tau 
    \right)
    \log 
    \sigma 
    \left( 
    \mathcal{S}(x_i,y_i), \tau 
    \right)
    \label{eq:kd_term}
\end{equation}
where $\hat{y}$ represents the output of $\mathcal{T}$, $\sigma$ denotes the softmax function with temperature $\tau$ and $\mathcal{S}$ represent the student network.

\subsubsection*{\textbf{Elastic Weight Consolidation (EWC)}}
\label{sec:ewc}
EWC remembers old tasks by selectively slowing down learning on weights 
that are important for these tasks. As shown in Kirkpatrick et al.~\cite{Kirkpatrick20173521}, learning from a task $A$ to a task $B$, there exist many configurations of $\theta$ leading to the same performance. In fact, the over-parametrization of the model makes it more likely the existence of a solution $\theta_B^*$ for task B that is close to task $A$. Therefore, previous tasks' performances are kept by constraining, with a quadratic penalty, the parameters to stay in a region centered in  $\theta_A^*$ of low error for task A. Formally, the function $\mathcal{L}$ that we minimize in EWC is:
\begin{equation}
    \mathcal{L}_{EWC}(\theta)=\mathcal{L}_B(\theta) + \sum_i \frac{\lambda}{2}F_i ( \theta_i - \theta^*_{A,i})^2
    \label{eq:ewc_loss}
\end{equation}
where $F$ is the Fisher information matrix, $\lambda$ sets how important the old task is compared to the new one and $i$ labels each parameter. When moving to a third task (i.e., task $C$), EWC will try to keep the network parameters 
close to the learned parameters of both task A and B. This can be enforced either with two separate penalties, or as one by noting that the sum of two quadratic penalties is itself a quadratic penalty. 

\subsection{Training procedure}
\label{sec:training}

The training procedure for both learning methods described in the previous section is summarized in Algorithm~\ref{alg:cl}. 
Given a stream of AI-generated contents $\mathcal{D}=\{D_1,\dots,D_n\}, n \geq 1$, at each training iteration $t$ we train a model $g_t(x_t, \theta_t)$ on the actual $\mathcal{D}_t \in \mathcal{D}$.
After the first training iteration on the first available batch of samples $\mathcal{D}_1 \in \mathcal{D}$, the model is trained on all next batches $\{\mathcal{D}_2,\dots,D_n\} \in \mathcal{D}$ for all $n > 1$. Differently from transfer learning, however, the model is forced to minimize the loss function for both new and old examples without requiring training over the past data samples, therefore learning an optimal set of parameters $\theta_t$ for all observed inputs $(x_i, y_i), i \in \{0,\dots,t\}$.

\begin{algorithm}
\caption{Training procedure for continual learning methods.}\label{alg:training}
\begin{algorithmic}
\Require $\mathcal{D}=\{\mathcal{D}_1,\dots, \mathcal{D}_n\}, n \geq 1$
\State $\mathcal{D}_t = \mathcal{D}.pop()$
\State $g_t(x_t, \theta_t) \gets train(\mathcal{D}_t)$

\While{$!\mathcal{D}.isEmpy()$}
    \State $\mathcal{D}_t = \mathcal{D}.pop()$
    
    \If{strategy == KD}
        \State $\mathcal{T} \gets g_{t}(x_t, \theta_t).copy()$
        \State $\mathcal{S} \gets train\left(\mathcal{T}, \mathcal{D}_t\right)$ 
        \Comment{Train using the $\mathcal{L}_{KD}$ loss in Eq. ~\ref{eq:kd_loss}.}
        \State $g_t(x_t, \theta_t) \gets \mathcal{S}$
    \ElsIf{strategy == EWC}
        \State $g_t(x_t, \theta_t) \gets train(\mathcal{D}_t)$ 
        \Comment{Train using the $\mathcal{L}_{EWC}$ loss in  Eq. ~\ref{eq:ewc_loss}.}
    \EndIf
\EndWhile
\end{algorithmic}
\label{alg:cl}
\end{algorithm}

\subsection{Lightweight CI/CD for deepfake detection}
\label{sec:cicd}

We conclude this section by providing an overview of how the proposed methodology could be integrated into a continuous integration and continuous delivery pipeline. The Figure~\ref{fig:mlops_pipeline} shows the complete pipeline. The upper part of the figure shows all the phases of an MLOps CI/CD system. The lower part shows the complete pipeline into which the continuous learning methods analyzed in this paper can be integrated. 

The pipeline consists of three main modules. In the first, data from generative tools, social media or databases are organized and analyzed if necessary by forensic experts or journalists. This module enables the preparation of model training data. In the next module, the model continuously learns from incoming data. This module also allows you to keep several copies of the model in case you need to restore a previous version. Finally, in the last part, a continous delivery and monitoring module takes care of serving the newly trained model to check for any data drift. If a new distribution of input data is identified (e.g., content generated with a new technique), it is immediately saved and flagged so that the model can be verified and possibly retrained on the new data.

\begin{figure*}
    \centering
    \includegraphics[width=0.7\textwidth]{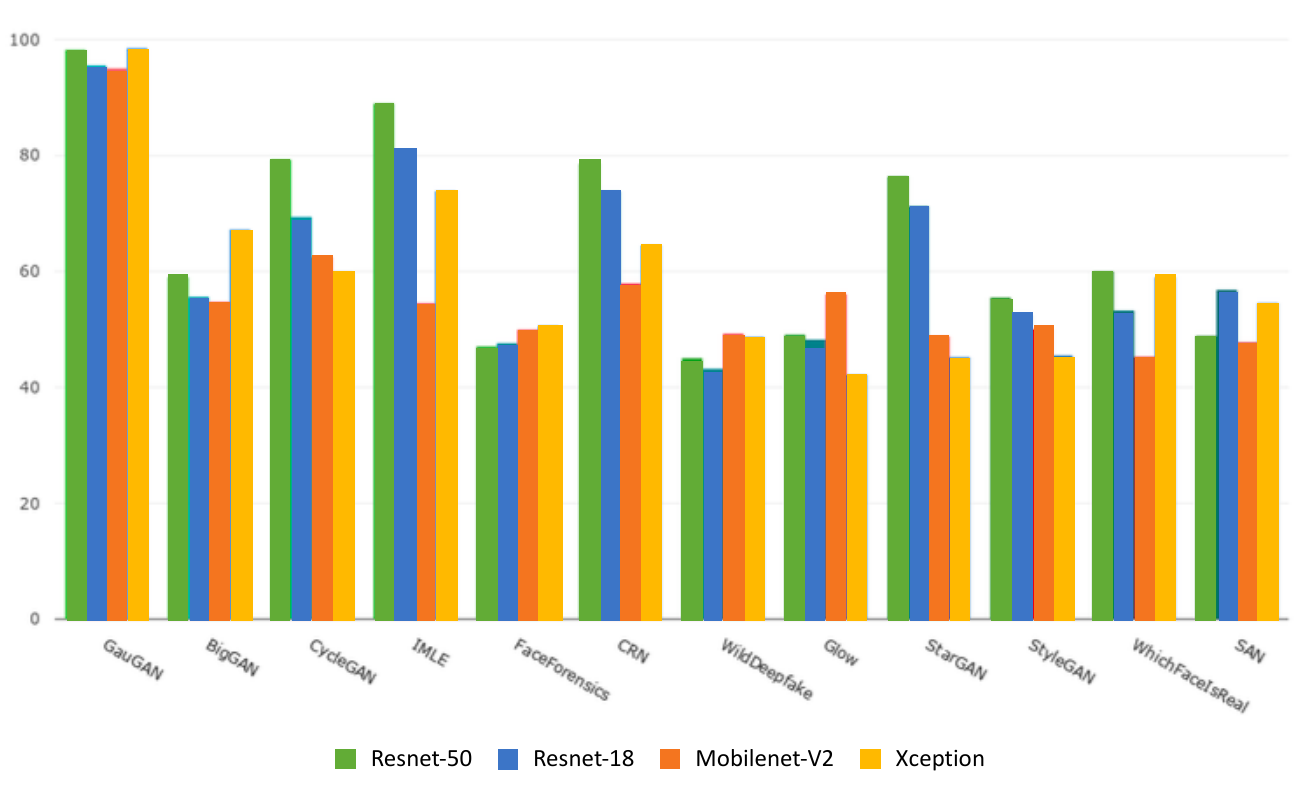}
    \caption{Zero-shot performance on the \emph{Long} set. All the models are trained only on the GauGAN task and evaluated over the whole dataset.}
    \label{fig:zeroshot}
\end{figure*}

\section{Experiments}
\label{sec:experiment}

In this section we discuss the analyses conducted to evaluate the effectiveness of the continual learning methods introduced in Section~\ref{sec:learning_strategies}. We begin by introducing the dataset and selected architectures in Section~\ref{sec:experimental_setting}, and then, in Section~\ref{sec:analysis} we analyze the performance of the two continuous learning strategies by comparing them with transfer learning.

\subsection{Experimental setting}

\label{sec:experimental_setting}
\subsubsection*{\textbf{Dataset}}
\label{sec:dataset}

We use the CDDB~\cite{Li20231339} dataset for all our experiments. The dataset offers three different evaluation
scenarios: an easy task sequence, a hard task sequence, and a long one. 
The dataset contains media generated with 5 different types of GAN-based generative models (StyleGAN, BigGAN, CycleGAN, GauGAN, and StarGAN), 5 non-GAN models (Glow, CRN, IMLE, SAN, and FaceForensics++), and two datasets whose origin is unknown (WhichFaceIsReal and WildDeepfake).

For this study, we report the results on the easy and long sequences. We refer to these two sequences as \emph{Easy} and \emph{Long}, respectively. 
\begin{itemize}
    \item The \emph{Easy} setup (composed of GauGAN, BigGAN, CycleGAN, IMLE, FaceForensic++, CRN, and WildDeepfake) is used to study the basic behavior of evaluated methods when they address similar generative techniques.
    \item The \emph{Long} setup (composed of GauGAN, BigGAN, CycleGAN, IMLE, FaceForensic++, CRN, WildDeepfake, Glow, StarGAN, StyleGAN, WhichFaceReal, and SAN) is designed to encourage methods to better handle long sequences of deepfake detection tasks, where the catastrophic forgetting might become more serious.
\end{itemize}

\subsubsection*{\textbf{Architecture}}
\label{sec:architecture}

For our analysis, we selected four different state-of-the-art that have demostrated to achieve good results on this task~\cite{gragnaniello2021gan,papa2023use}: Resnet-50, Resnet-18, Mobilenet-V2, and Xception. 

\subsection{Analysis}
\label{sec:analysis}

We now turn to analyze the performance of continuous learning methods with selected backbones. We start by illustrating the results on the \emph{short} set and then extend the considerations to the longer (i.e., the \emph{long} set). We trained each model using early stopping with a patience of 35 up to 250 epochs. We used the Stochastic Gradient Descent (SGD) optimizer to modify the models’ weights, with an initial learning rate of 0.005 and a momentum of 0.1. The learning rate is controlled by a cosine annealing scheduler with a minimum value of $10^{-5}$. For preprocessing, we applied a random cropped on each image with a resolution of $128\times128$.

\subsubsection*{\textbf{Zero shot}}
\label{sec:zeroshot}

We begin by analyzing the results of all models trained on the GauGAN task and evaluating the whole \emph{long} sequence. As shown in Figure~\ref{fig:zeroshot}, all the models almost always fail to detect tasks outside their training data, indicating a clear lack of generalizability. These evaluations confirm that a model trained on a particular generative technique struggles to detect other types of fake images. In fact, we can see that the model manages to achieve more or less satisfactory performances on media generated with GANs (in particular BigGAN, CycleGAN, and StarGAN), which evidently have characteristics more similar to those seen in the training phase, but it ultimately fails the tasks more complex ones like FaceForensics++ or WildDeepfake.

\subsubsection*{\textbf{Short set}}
\label{sec:shortset}

In Table~\ref{tab:easy}, we report the experiments conducted on this set. The table reports the performance of each dataset trained on the entire sequence. For example, Resnet-50 obtains an accuracy of $57.80\%$ on GauGAN after being trained with KD up to the last available dataset in the sequence (i.e., WildDeepfake). From the table, we can draw some initial insights. Starting with the backbones, the Resnet-50 and Mobilenet-V2 achieve the best results on average. As for learning techniques, we can see a general improvement in the performance of continual learning methods compared to transfer learning. In particular, Knowledge Distillation achieves the best performance when combined with Resnet-50. In this specific configuration, the model achieves excellent performance on IMLE (97.30\%) and CRN (94.36\%), and good performance on CycleGAN (79.85\%), while the results on the other datasets range between 51.22\% and 64.45\%. The results are not surprisingly high, but to understand why, it is necessary to reason about the different types of datasets contained in this sequence. IMLE and CRN both contain media drawn from computer games, which are indeed more easily recognized. GauGAN, BigGAN, and CycleGAN are all datasets generated with Generative Adversarial Networks, so they have similar characteristics. Finally, FaceForensics++ and WildDeepfake are both challenging sets containing diverse generative techniques. These characteristics make the sequence highly varied and therefore complex to be classified uniformly well. In particular, as we iterated through the various tasks in the sequence, we noticed that performance can fluctuate significantly depending on the order in which these datasets are used. 

\begin{figure*}
    \centering
    \includegraphics[width=0.7\linewidth]{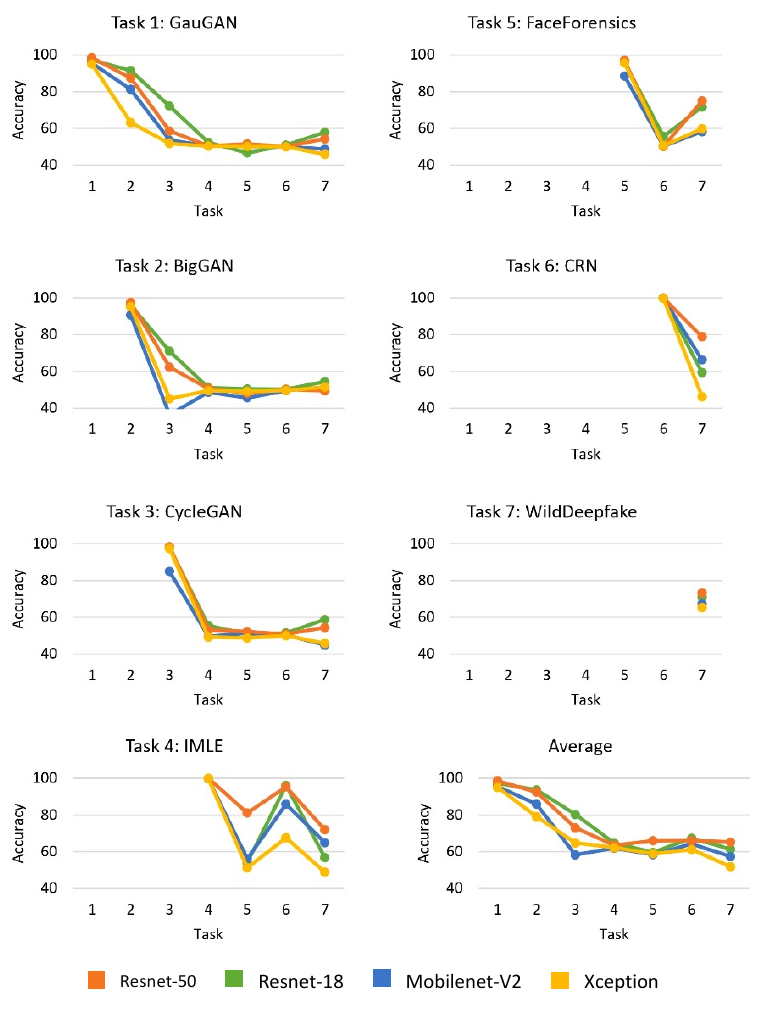}
    \caption{Elastic Weight Consolidation average accuracy at each task $t$ calculated over tasks $\{1,\dots,t\}$. of all backbones on the \emph{Easy} set. The order of the tasks is the following: (1) GauGAN, (2) BigGAN, (3) CycleGAN, (4) IMLE, (5) FaceForensic++, (6) CRN, and (7) WildDeepfake.}
    \label{fig:ewc}
\end{figure*}

\begin{figure*}
    \centering
    \includegraphics[width=0.7\linewidth]{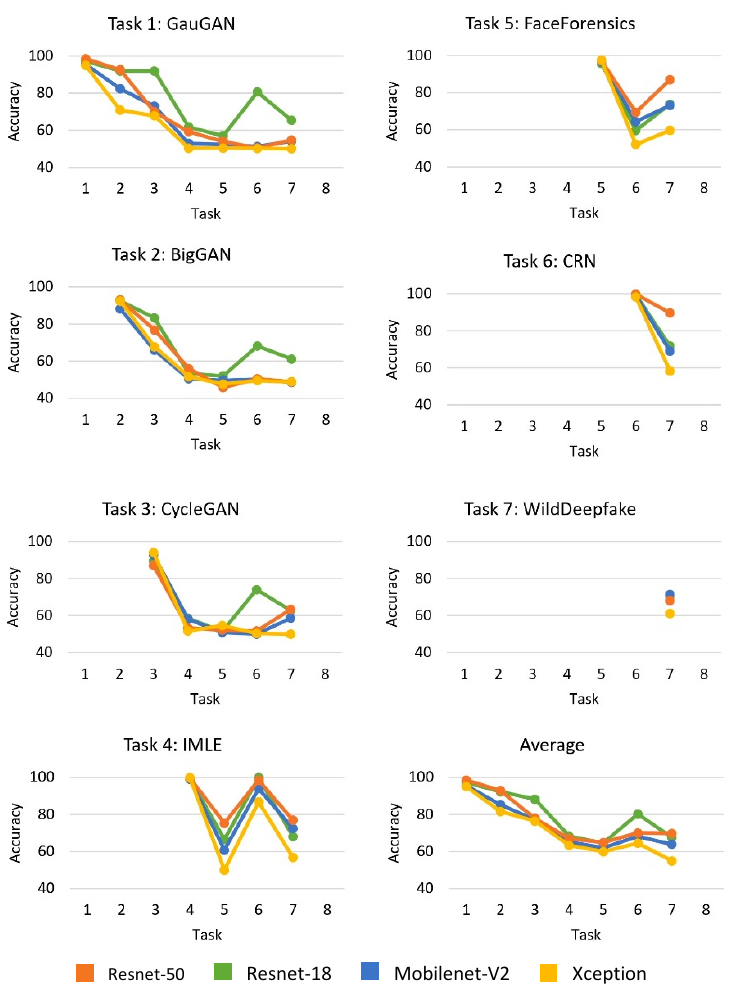}
    \caption{Knowledge distillation average accuracy at each task $t$ calculated over tasks $\{1,\dots,t\}$. of all backbones on the \emph{Easy} set. The order of the tasks is the following: (1) GauGAN, (2) BigGAN, (3) CycleGAN, (4) IMLE, (5) FaceForensic++, (6) CRN, and (7) WildDeepfake.}
    \label{fig:kd}
\end{figure*}

\begin{table*}
    \centering
    \resizebox{\textwidth}{!}{
        \begin{tabular}{|c|c|ccccccc|c|}
            \hline
            \multirow{2}{*}{Method} & \multirow{2}{*}{Model} & \multicolumn{7}{c|}{Type} & \multirow{2}{*}{Average}\\
            \cline{3-9}
            &  & GauGAN & BigGAN & CycleGAN & IMLE & FaceForensics++ & CRN & WildDeepfake & \\
            \hline
            \multirow{4}{*}{Transfer learning} & ResNet-50 & \textbf{57.80} & \textbf{54.38} & \textbf{58.79} & \underline{56.89} & \underline{71.63} & 59.32 & \underline{71.02} & \underline{61.40} \\
                                               & ResNet-18 & 48.70 & 51.50 & 44.87 & 65.02 & 58.23 & \underline{66.25} & 67.32 & 57.41 \\
                                               & Mobilenet-V2 & \underline{54.30} & \underline{49.38} & \underline{54.40} & \textbf{72.03} & \textbf{74.88} & \textbf{78.82}  & \textbf{73.31} & \textbf{65.30} \\
                                               & Xception & 45.90 & 51.38 & 45.79 & 48.87 & 59.81 & 46.16 & 65.26 & 51.88 \\
            \hline
            \multirow{4}{*}{EWC} & ResNet-50 & \textbf{65.30} & \textbf{61.13} & \underline{62.64} & 68.11 & \underline{73.77} & \underline{71.77} & \textbf{70.24} & \underline{67.56} \\
                                 & ResNet-18 & 54.15 & 48.50 & 58.61 & \underline{72.50} & 73.21 & 69.07 & 71.36 & 63.91 \\
                                 & Mobilenet-V2 & \underline{54.75} & \underline{48.88} & \textbf{63.37} & \textbf{76.96} & \textbf{86.98} & \textbf{89.62} & \underline{68.03} & \textbf{69.80} \\
                                 & Xception & 50.00 & 48.75 & 50.00 & 56.85 & 59.72 & 58.30 & 61.06 & 54.95 \\
            \hline
            \multirow{4}{*}{KD} & ResNet-50 & \textbf{64.45} & \textbf{58.75} & \textbf{79.85} & \underline{97.30} & \underline{64.65} & \underline{94.36} & \textbf{51.22} & \textbf{72.94} \\
                                & ResNet-18 & 50.50 & 49.13 & 52.75 & 52.27 & 52.28 & 53.45 & 49.87 & 51.46 \\
                                & Mobilenet-V2 & \underline{54.40} & \underline{53.00} & \underline{54.21} & 80.28 & \textbf{72.19} & 86.37 & \underline{71.36} & \underline{67.40} \\
                                & Xception & 50.75 & 50.75 & 50.55 & \textbf{97.73} & 49.86 & \textbf{96.28} & 49.57 & 63.64 \\
            \hline
        \end{tabular}
    }
    \caption{Transfer learning, EWC, and KD performance on the \emph{easy} set. The average column represents the average performance of each model across all datasets. Bold values represent the highest value for each learning method.}
    \label{tab:easy}
\end{table*}

In Figures~\ref{fig:ewc} and~\ref{fig:kd}, we elaborate more on this behavior. The figure shows the average performance on each dataset recorded by training the model on the various tasks in the following order: (1) GauGAN, (2) BigGAN, (3) CycleGAN, (4) IMLE, (5) FaceForensic++, (6) CRN, and (7) WildDeepfake. From the figure, we can see that the performance remains high on average on the first three (GAN-based) tasks and then undergoes an initial slight decrease with IMLE and stabilizes with a more substantial decrease from the fifth task onward. By analyzing this behavior, we can infer that the big difference between the generative techniques present in FaceForensic++ and the previous tasks strains the model in finding a region of optimum that reduces the error on all tasks. This result is further confirmed with the arrival of WildDeepfake, which turns out to be the most complex and different task overall than the previous ones. The strong \emph{asymmetry} of some tasks compared to others seems to play an important role, leading the model to optimize performance towards some families of tasks rather than others.

\begin{figure}
    \centering
    \begin{subfigure}{0.45\textwidth}
        \centering
        \includegraphics[width=0.8\textwidth]{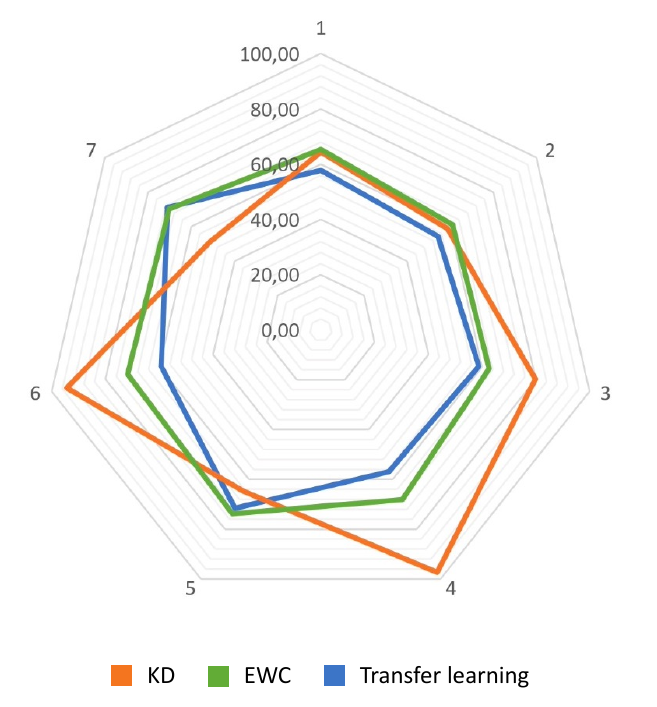}
        \caption{Complete \emph{Easy} set.}
        \label{fig:short}
    \end{subfigure}
    \begin{subfigure}{0.45\textwidth}
        \centering
        \includegraphics[width=0.8\textwidth]{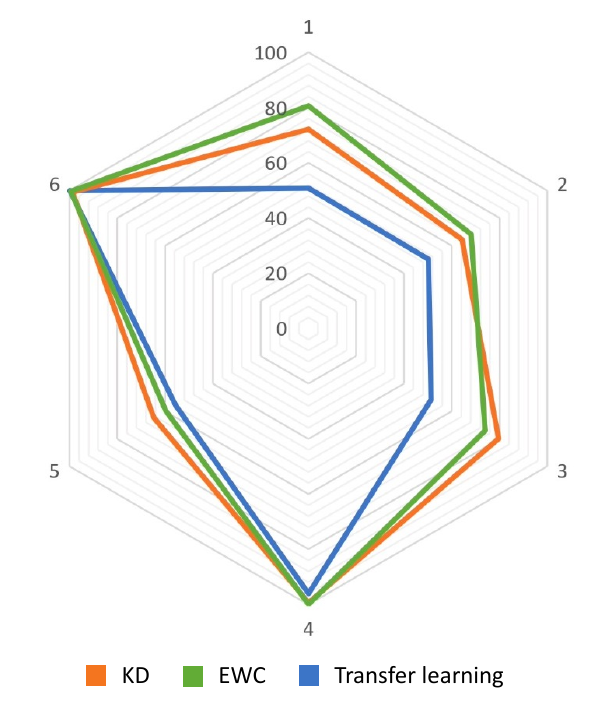}
        \caption{\emph{Easy} set without WildDeepfake.}
        \label{fig:short-wd}
    \end{subfigure}
    \caption{The average accuracy of Resnet-50 trained with KD on the full \emph{Easy} set (Figure~\ref{fig:short}) and without WildDeepfake (Figure~\ref{fig:short-wd}). Some datasets seem to heavily afflict performance in the continuous learning context.}
    \label{fig:short-easy}
\end{figure}

To confirm this hypothesis we tried to repeat the experiment by removing WildDeepfake from the sequence. As we can see from Figure~\ref{fig:short-easy}, the performance of the models improves significantly if we compare the performance of the model trained on the complete sequence (Figure~\ref{fig:short}) compared to the one trained on the sequence without WildDeepfake (Figure~\ref{fig:short-wd}). This tells us that the symmetry between the different tasks is fundamental to maintaining overall satisfactory performance on the entire sequence.

Overall, the results suggest a few things. First, we can notice a greater stability of continuous learning techniques compared to transfer learning. Furthermore, Knowledge Distillation appears to be more robust overall than EWC, achieving higher performance on average than its rival. In all cases, all techniques achieve better results than the zero-shot scenario. However, the order of the sequence and the similarity of the tasks seem to play a significant role in overall performance. Finally, the most robust model seems to be the Resnet-50, followed closely by the Mobilenet-V2.

\subsubsection*{\textbf{Long set}}
\label{sec:longset}

\begin{figure*}
    \centering
    \includegraphics[width=0.6\textwidth]{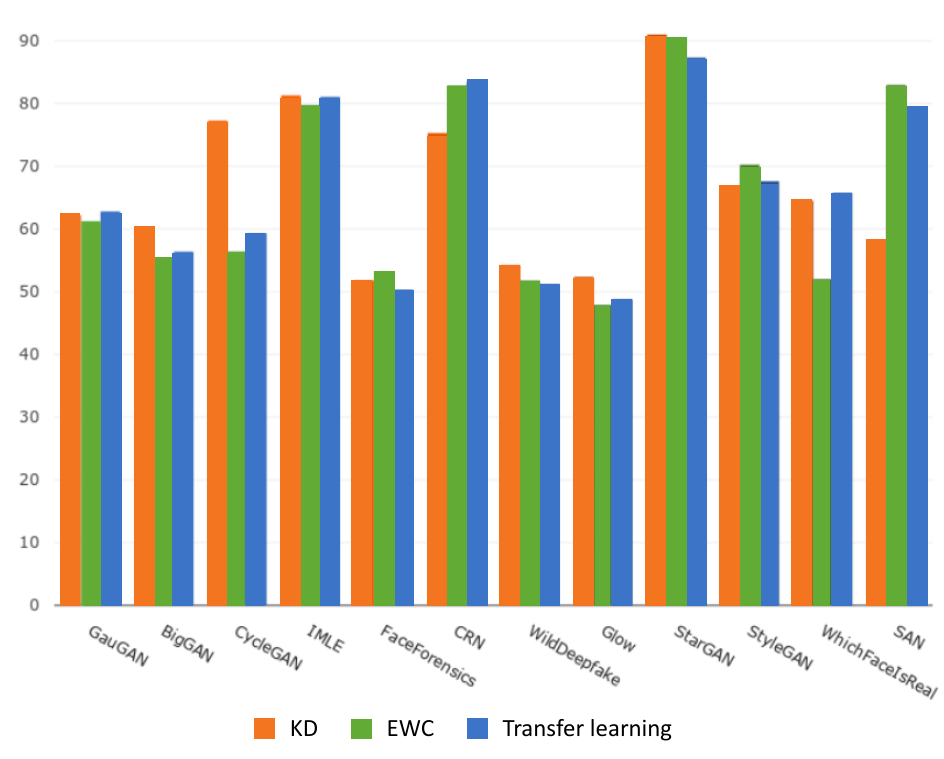}
    \caption{Average accuracy for Resnet-50 on the \emph{Long} set.}
    \label{fig:long}
\end{figure*}

To complete the analysis, we extend the considerations made up to now to the \emph{long} sequence. In Figure~\ref{fig:long}, we report the average accuracy values on each task. As for the \emph{short} sequence, the heterogeneity of the 12 tasks poses a challenge to the continual learning methods, preventing them from learning a general representation applicable to all tasks. In particular, the similar result between the continual learning methods and transfer learning suggests that the latter cannot mitigate \emph{catastrophic forgetting}. In fact, we can note that the models are able to achieve good performance on some tasks like IMLE and CRN (which have similar characteristics) and GAN-based tasks like GauGAN, BigGAN, StarGAN, and StyleGAN but fail to maintain acceptable performance on the other tasks.

\subsubsection*{\textbf{Multi-task sequences}}
\label{sec:multitask}

\begin{figure*}
    \centering
    \includegraphics[width=0.6\textwidth]{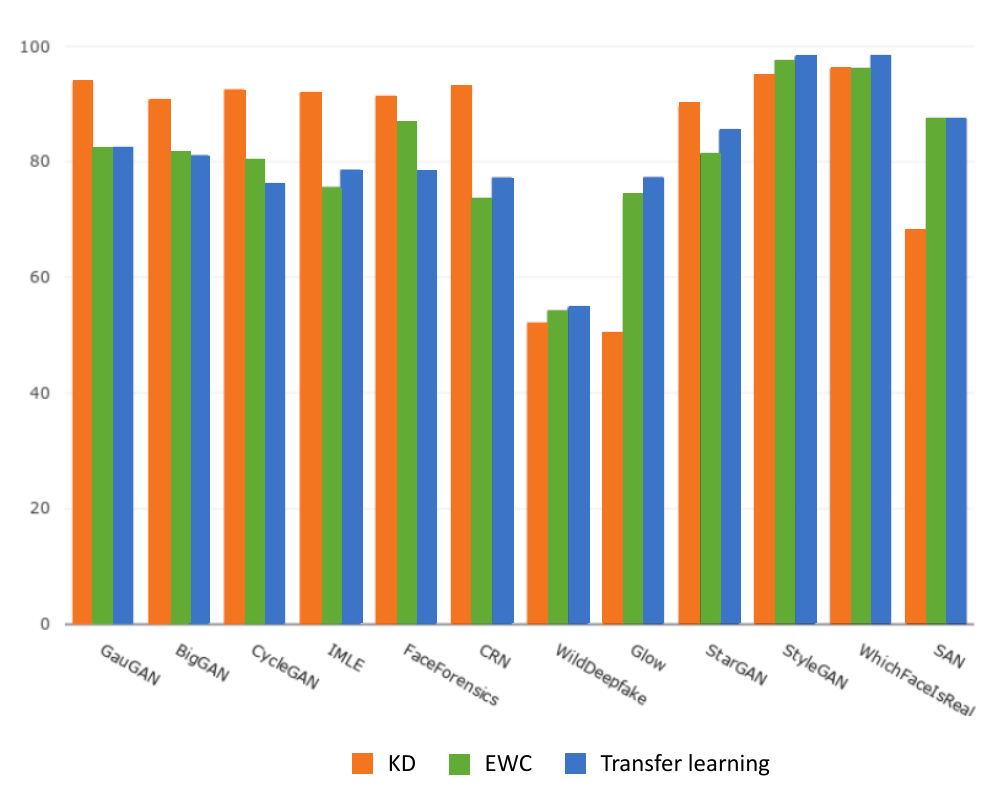}
    \caption{Average accuracy for Resnet-50 trained with the multi-task configuration on the \emph{Long} set.}
    \label{fig:multitask}
\end{figure*}

Collecting the considerations made so far, in this last test, we test the methods on a hybrid continuous learning scenario. Instead of learning each individual task separately, we combined the \emph{Long} sequence tasks into groups of three, which we call \emph{Multi-tasks}. Therefore, in this configuration, the sequence is composed of 4 macro tasks: $t1 = \{$\emph{GauGAN, BigGAN, CycleGAN}\}, $t2 = \{$\emph{IMLE, FaceForensics, CRN}$\}$, $t3 = \{ $\emph{WildDeepfake, Glow, StarGAN}$\}$, and $t4 = \{$\emph{StyleGAN, WhichFaceReal, SAN}$\}$. Figure~\ref{fig:multitask} reports the average accuracy values on each task. From the figure, we can immediately see that, as suggested in the previous experiment, WildDeepfake puts a strain on all learning techniques. It becomes clear that including images from unknown sources and tasks with a restricted dataset causes a drastic drop in accuracy. Moreover, regarding the learning techniques, all methods achieved satisfactory results. In particular, Knowledge Distillation confirms itself as more robust in most of the sequence but seems to suffer a slight decline in the last three tasks. This result confirms that by aggregating tasks on the basis of their similarity, continuous learning techniques are able to maintain good performance over time.

\subsection{Limitations}
\label{sec:limitations}

The analyzed results give us some interesting information. First of all, continuous learning techniques show an improvement compared to the zero-shot scenario. However, the similarity of the various tasks in sequence and their order of arrival seem to play an important role in maintaining satisfactory performance throughout the sequence.

The results presented require further investigation. First of all, in this study, we limited ourselves to analyzing the behavior of 4 backbones commonly used in deepfake detection; however, a comparison of more advanced deepfake detection methods present in the literature is necessary. Furthermore, the results obtained in the multi-task configuration suggest that these techniques could benefit from using a memory~\cite{del2023studying}. This aspect will be further analyzed in our future studies. Finally, in this study, we assumed the different tasks arrived in batches. This is certainly possible, as the release of a new tool could lead to introducing a significant sample of media generated with it. However, it remains essential to study the robustness of these techniques on smaller and more heterogeneous sequences.

\section{Conclusion}
\label{sec:conclusion}

In this paper, we proposed an analysis of two deepfake detection techniques (Knowledge Distillation and Elastic Weight Consolidation), comparing them with zero-shot scenarios and transfer learning. The results show that continuous learning techniques can help make deepfake detection models more robust and easily updated to new generative methods. However, our analysis also highlighted some problems. The performance of these learning strategies seems to depend significantly on the similarity of the tasks and their order of arrival. To address this problem, we have shown how it is possible to combine the different tasks to obtain significantly better performance. Consequently, future developments of this work could analyze the importance of using memory and the robustness of these learning techniques to smaller and more heterogeneous sequences.

In addition to this, we also gave an overview of a CI/CD pipeline for deepfake detection, showing how the models used in this work can be combined with other modules to obtain a pipeline that can be used in a real application scenario. In this regard, in the future we will present a complete version of all the fundamental modules of this pipeline, starting first from the data drift detection module. This module, in particular, could help to significantly improve performance in continuous learning.

\section*{Acknowledgments}
This study has been partially supported by SERICS (PE00000014) under the MUR National Recovery and Resilience Plan funded by the European Union - NextGenerationEU and Sapienza University of Rome project “EV2” (003\_009\_22).

\bibliographystyle{ACM-Reference-Format}
\bibliography{main}


\begin{thebibliography}{32}


\ifx \showCODEN    \undefined \def \showCODEN     #1{\unskip}     \fi
\ifx \showDOI      \undefined \def \showDOI       #1{#1}\fi
\ifx \showISBNx    \undefined \def \showISBNx     #1{\unskip}     \fi
\ifx \showISBNxiii \undefined \def \showISBNxiii  #1{\unskip}     \fi
\ifx \showISSN     \undefined \def \showISSN      #1{\unskip}     \fi
\ifx \showLCCN     \undefined \def \showLCCN      #1{\unskip}     \fi
\ifx \shownote     \undefined \def \shownote      #1{#1}          \fi
\ifx \showarticletitle \undefined \def \showarticletitle #1{#1}   \fi
\ifx \showURL      \undefined \def \showURL       {\relax}        \fi
\providecommand\bibfield[2]{#2}
\providecommand\bibinfo[2]{#2}
\providecommand\natexlab[1]{#1}
\providecommand\showeprint[2][]{arXiv:#2}

\bibitem[Aghasanli et~al\mbox{.}(2023)]%
        {aghasanli2023interpretable}
\bibfield{author}{\bibinfo{person}{Agil Aghasanli}, \bibinfo{person}{Dmitry Kangin}, {and} \bibinfo{person}{Plamen Angelov}.} \bibinfo{year}{2023}\natexlab{}.
\newblock \showarticletitle{Interpretable-through-prototypes deepfake detection for diffusion models}. In \bibinfo{booktitle}{\emph{Proceedings of the IEEE/CVF International Conference on Computer Vision}}. \bibinfo{pages}{467--474}.
\newblock


\bibitem[Amerini et~al\mbox{.}(2021)]%
        {Amerini2021309}
\bibfield{author}{\bibinfo{person}{Irene Amerini}, \bibinfo{person}{Aris Anagnostopoulos}, \bibinfo{person}{Luca Maiano}, {and} \bibinfo{person}{Lorenzo~Ricciardi Celsi}.} \bibinfo{year}{2021}\natexlab{}.
\newblock \showarticletitle{Deep learning for multimedia forensics}.
\newblock \bibinfo{journal}{\emph{Foundations and Trends in Computer Graphics and Vision}} \bibinfo{volume}{12}, \bibinfo{number}{4} (\bibinfo{year}{2021}), \bibinfo{pages}{309 – 457}.
\newblock
\urldef\tempurl%
\url{https://doi.org/10.1561/0600000096}
\showDOI{\tempurl}


\bibitem[Chai et~al\mbox{.}(2020)]%
        {chai2020makes}
\bibfield{author}{\bibinfo{person}{Lucy Chai}, \bibinfo{person}{David Bau}, \bibinfo{person}{Ser-Nam Lim}, {and} \bibinfo{person}{Phillip Isola}.} \bibinfo{year}{2020}\natexlab{}.
\newblock \showarticletitle{What makes fake images detectable? understanding properties that generalize}. In \bibinfo{booktitle}{\emph{Computer Vision--ECCV 2020: 16th European Conference, Glasgow, UK, August 23--28, 2020, Proceedings, Part XXVI 16}}. Springer, \bibinfo{pages}{103--120}.
\newblock


\bibitem[Corvi et~al\mbox{.}(2023)]%
        {corvi2023intriguing}
\bibfield{author}{\bibinfo{person}{Riccardo Corvi}, \bibinfo{person}{Davide Cozzolino}, \bibinfo{person}{Giovanni Poggi}, \bibinfo{person}{Koki Nagano}, {and} \bibinfo{person}{Luisa Verdoliva}.} \bibinfo{year}{2023}\natexlab{}.
\newblock \showarticletitle{Intriguing properties of synthetic images: from generative adversarial networks to diffusion models}. In \bibinfo{booktitle}{\emph{Proceedings of the IEEE/CVF Conference on Computer Vision and Pattern Recognition}}. \bibinfo{pages}{973--982}.
\newblock


\bibitem[De~Lange et~al\mbox{.}(2022)]%
        {DeLange20223366}
\bibfield{author}{\bibinfo{person}{Matthias De~Lange}, \bibinfo{person}{Rahaf Aljundi}, \bibinfo{person}{Marc Masana}, \bibinfo{person}{Sarah Parisot}, \bibinfo{person}{Xu Jia}, \bibinfo{person}{Ales Leonardis}, \bibinfo{person}{Gregory Slabaugh}, {and} \bibinfo{person}{Tinne Tuytelaars}.} \bibinfo{year}{2022}\natexlab{}.
\newblock \showarticletitle{A Continual Learning Survey: Defying Forgetting in Classification Tasks}.
\newblock \bibinfo{journal}{\emph{IEEE Transactions on Pattern Analysis and Machine Intelligence}} \bibinfo{volume}{44}, \bibinfo{number}{7} (\bibinfo{year}{2022}), \bibinfo{pages}{3366 – 3385}.
\newblock
\urldef\tempurl%
\url{https://doi.org/10.1109/TPAMI.2021.3057446}
\showDOI{\tempurl}


\bibitem[del Rio et~al\mbox{.}(2023)]%
        {del2023studying}
\bibfield{author}{\bibinfo{person}{Felipe del Rio}, \bibinfo{person}{Julio Hurtado}, \bibinfo{person}{Cristian Buc}, \bibinfo{person}{Alvaro Soto}, {and} \bibinfo{person}{Vincenzo Lomonaco}.} \bibinfo{year}{2023}\natexlab{}.
\newblock \showarticletitle{Studying Generalization on Memory-Based Methods in Continual Learning}.
\newblock \bibinfo{journal}{\emph{arXiv preprint arXiv:2306.09890}} (\bibinfo{year}{2023}).
\newblock


\bibitem[Fei et~al\mbox{.}(2023)]%
        {fei2023robust}
\bibfield{author}{\bibinfo{person}{Jianwei Fei}, \bibinfo{person}{Zhihua Xia}, \bibinfo{person}{Benedetta Tondi}, {and} \bibinfo{person}{Mauro Barni}.} \bibinfo{year}{2023}\natexlab{}.
\newblock \showarticletitle{Robust Retraining-free GAN Fingerprinting via Personalized Normalization}. In \bibinfo{booktitle}{\emph{2023 IEEE International Workshop on Information Forensics and Security (WIFS)}}. IEEE, \bibinfo{pages}{1--6}.
\newblock


\bibitem[French(1999)]%
        {French1999128}
\bibfield{author}{\bibinfo{person}{Robert~M. French}.} \bibinfo{year}{1999}\natexlab{}.
\newblock \showarticletitle{Catastrophic forgetting in connectionist networks}.
\newblock \bibinfo{journal}{\emph{Trends in Cognitive Sciences}} \bibinfo{volume}{3}, \bibinfo{number}{4} (\bibinfo{year}{1999}), \bibinfo{pages}{128 – 135}.
\newblock
\urldef\tempurl%
\url{https://doi.org/10.1016/S1364-6613(99)01294-2}
\showDOI{\tempurl}


\bibitem[Gragnaniello et~al\mbox{.}(2021)]%
        {gragnaniello2021gan}
\bibfield{author}{\bibinfo{person}{Diego Gragnaniello}, \bibinfo{person}{Davide Cozzolino}, \bibinfo{person}{Francesco Marra}, \bibinfo{person}{Giovanni Poggi}, {and} \bibinfo{person}{Luisa Verdoliva}.} \bibinfo{year}{2021}\natexlab{}.
\newblock \showarticletitle{Are GAN generated images easy to detect? A critical analysis of the state-of-the-art}. In \bibinfo{booktitle}{\emph{2021 IEEE international conference on multimedia and expo (ICME)}}. IEEE, \bibinfo{pages}{1--6}.
\newblock


\bibitem[Hinton et~al\mbox{.}(2015)]%
        {hinton2015distilling}
\bibfield{author}{\bibinfo{person}{Geoffrey Hinton}, \bibinfo{person}{Oriol Vinyals}, {and} \bibinfo{person}{Jeff Dean}.} \bibinfo{year}{2015}\natexlab{}.
\newblock \showarticletitle{Distilling the knowledge in a neural network}.
\newblock \bibinfo{journal}{\emph{arXiv preprint arXiv:1503.02531}} (\bibinfo{year}{2015}).
\newblock


\bibitem[Ju et~al\mbox{.}(2022)]%
        {ju2022fusing}
\bibfield{author}{\bibinfo{person}{Yan Ju}, \bibinfo{person}{Shan Jia}, \bibinfo{person}{Lipeng Ke}, \bibinfo{person}{Hongfei Xue}, \bibinfo{person}{Koki Nagano}, {and} \bibinfo{person}{Siwei Lyu}.} \bibinfo{year}{2022}\natexlab{}.
\newblock \showarticletitle{Fusing global and local features for generalized ai-synthesized image detection}. In \bibinfo{booktitle}{\emph{2022 IEEE International Conference on Image Processing (ICIP)}}. IEEE, \bibinfo{pages}{3465--3469}.
\newblock


\bibitem[Khan and Dai(2021)]%
        {khan2021video}
\bibfield{author}{\bibinfo{person}{Sohail~Ahmed Khan} {and} \bibinfo{person}{Hang Dai}.} \bibinfo{year}{2021}\natexlab{}.
\newblock \showarticletitle{Video transformer for deepfake detection with incremental learning}. In \bibinfo{booktitle}{\emph{Proceedings of the 29th ACM International Conference on Multimedia}}. \bibinfo{pages}{1821--1828}.
\newblock


\bibitem[Kim et~al\mbox{.}(2021)]%
        {kim2021cored}
\bibfield{author}{\bibinfo{person}{Minha Kim}, \bibinfo{person}{Shahroz Tariq}, {and} \bibinfo{person}{Simon~S Woo}.} \bibinfo{year}{2021}\natexlab{}.
\newblock \showarticletitle{Cored: Generalizing fake media detection with continual representation using distillation}. In \bibinfo{booktitle}{\emph{Proceedings of the 29th ACM International Conference on Multimedia}}. \bibinfo{pages}{337--346}.
\newblock


\bibitem[Kirkpatrick et~al\mbox{.}(2017)]%
        {Kirkpatrick20173521}
\bibfield{author}{\bibinfo{person}{James Kirkpatrick}, \bibinfo{person}{Razvan Pascanu}, \bibinfo{person}{Neil Rabinowitz}, \bibinfo{person}{Joel Veness}, \bibinfo{person}{Guillaume Desjardins}, \bibinfo{person}{Andrei~A Rusu}, \bibinfo{person}{Kieran Milan}, \bibinfo{person}{John Quan}, \bibinfo{person}{Tiago Ramalho}, \bibinfo{person}{Agnieszka Grabska-Barwinska}, \bibinfo{person}{Demis Hassabis}, \bibinfo{person}{Claudia Clopath}, \bibinfo{person}{Dharshan Kumaran}, {and} \bibinfo{person}{Raia Hadsell}.} \bibinfo{year}{2017}\natexlab{}.
\newblock \showarticletitle{Overcoming catastrophic forgetting in neural networks}.
\newblock \bibinfo{journal}{\emph{Proceedings of the National Academy of Sciences of the United States of America}} \bibinfo{volume}{114}, \bibinfo{number}{13} (\bibinfo{year}{2017}), \bibinfo{pages}{3521 – 3526}.
\newblock
\urldef\tempurl%
\url{https://doi.org/10.1073/pnas.1611835114}
\showDOI{\tempurl}


\bibitem[Le and Woo(2023)]%
        {le2023quality}
\bibfield{author}{\bibinfo{person}{Binh~M Le} {and} \bibinfo{person}{Simon~S Woo}.} \bibinfo{year}{2023}\natexlab{}.
\newblock \showarticletitle{Quality-agnostic deepfake detection with intra-model collaborative learning}. In \bibinfo{booktitle}{\emph{Proceedings of the IEEE/CVF International Conference on Computer Vision}}. \bibinfo{pages}{22378--22389}.
\newblock


\bibitem[Li et~al\mbox{.}(2023)]%
        {Li20231339}
\bibfield{author}{\bibinfo{person}{Chuqiao Li}, \bibinfo{person}{Zhiwu Huang}, \bibinfo{person}{Danda~Pani Paudel}, \bibinfo{person}{Yabin Wang}, \bibinfo{person}{Mohamad Shahbazi}, \bibinfo{person}{Xiaopeng Hong}, {and} \bibinfo{person}{Luc Van~Gool}.} \bibinfo{year}{2023}\natexlab{}.
\newblock \showarticletitle{A Continual Deepfake Detection Benchmark: Dataset, Methods, and Essentials}.
\newblock \bibinfo{journal}{\emph{Proceedings - 2023 IEEE Winter Conference on Applications of Computer Vision, WACV 2023}} (\bibinfo{year}{2023}), \bibinfo{pages}{1339 – 1349}.
\newblock
\urldef\tempurl%
\url{https://doi.org/10.1109/WACV56688.2023.00139}
\showDOI{\tempurl}


\bibitem[Maiano et~al\mbox{.}(2022)]%
        {maiano2022depthfake}
\bibfield{author}{\bibinfo{person}{Luca Maiano}, \bibinfo{person}{Lorenzo Papa}, \bibinfo{person}{Ketbjano Vocaj}, {and} \bibinfo{person}{Irene Amerini}.} \bibinfo{year}{2022}\natexlab{}.
\newblock \showarticletitle{DepthFake: a depth-based strategy for detecting Deepfake videos}.
\newblock \bibinfo{journal}{\emph{arXiv preprint arXiv:2208.11074}} (\bibinfo{year}{2022}).
\newblock


\bibitem[Marra et~al\mbox{.}(2019)]%
        {marra2019incremental}
\bibfield{author}{\bibinfo{person}{Francesco Marra}, \bibinfo{person}{Cristiano Saltori}, \bibinfo{person}{Giulia Boato}, {and} \bibinfo{person}{Luisa Verdoliva}.} \bibinfo{year}{2019}\natexlab{}.
\newblock \showarticletitle{Incremental learning for the detection and classification of gan-generated images}. In \bibinfo{booktitle}{\emph{2019 IEEE international workshop on information forensics and security (WIFS)}}. IEEE, \bibinfo{pages}{1--6}.
\newblock


\bibitem[Paleyes et~al\mbox{.}(2022)]%
        {Paleyes2022}
\bibfield{author}{\bibinfo{person}{Andrei Paleyes}, \bibinfo{person}{Raoul-Gabriel Urma}, {and} \bibinfo{person}{Neil~D. Lawrence}.} \bibinfo{year}{2022}\natexlab{}.
\newblock \showarticletitle{Challenges in Deploying Machine Learning: A Survey of Case Studies}.
\newblock \bibinfo{journal}{\emph{Comput. Surveys}} \bibinfo{volume}{55}, \bibinfo{number}{6} (\bibinfo{year}{2022}).
\newblock
\urldef\tempurl%
\url{https://doi.org/10.1145/3533378}
\showDOI{\tempurl}


\bibitem[Pan et~al\mbox{.}(2023)]%
        {Pan20238035}
\bibfield{author}{\bibinfo{person}{Kun Pan}, \bibinfo{person}{Yifang Yin}, \bibinfo{person}{Yao Wei}, \bibinfo{person}{Feng Lin}, \bibinfo{person}{Zhongjie Ba}, \bibinfo{person}{Zhenguang Liu}, \bibinfo{person}{Zhibo Wang}, \bibinfo{person}{Lorenzo Cavallaro}, {and} \bibinfo{person}{Kui Ren}.} \bibinfo{year}{2023}\natexlab{}.
\newblock \showarticletitle{DFIL: Deepfake Incremental Learning by Exploiting Domain-invariant Forgery Clues}.
\newblock \bibinfo{journal}{\emph{MM 2023 - Proceedings of the 31st ACM International Conference on Multimedia}} (\bibinfo{year}{2023}), \bibinfo{pages}{8035 – 8046}.
\newblock
\urldef\tempurl%
\url{https://doi.org/10.1145/3581783.3612377}
\showDOI{\tempurl}


\bibitem[Papa et~al\mbox{.}(2023)]%
        {papa2023use}
\bibfield{author}{\bibinfo{person}{Lorenzo Papa}, \bibinfo{person}{Lorenzo Faiella}, \bibinfo{person}{Luca Corvitto}, \bibinfo{person}{Luca Maiano}, {and} \bibinfo{person}{Irene Amerini}.} \bibinfo{year}{2023}\natexlab{}.
\newblock \showarticletitle{On the use of Stable Diffusion for creating realistic faces: from generation to detection}. In \bibinfo{booktitle}{\emph{2023 11th International Workshop on Biometrics and Forensics (IWBF)}}. IEEE, \bibinfo{pages}{1--6}.
\newblock


\bibitem[Raza and Malik(2023)]%
        {raza2023multimodaltrace}
\bibfield{author}{\bibinfo{person}{Muhammad~Anas Raza} {and} \bibinfo{person}{Khalid~Mahmood Malik}.} \bibinfo{year}{2023}\natexlab{}.
\newblock \showarticletitle{Multimodaltrace: Deepfake Detection Using Audiovisual Representation Learning}. In \bibinfo{booktitle}{\emph{Proceedings of the IEEE/CVF Conference on Computer Vision and Pattern Recognition}}. \bibinfo{pages}{993--1000}.
\newblock


\bibitem[Rebuffi et~al\mbox{.}(2017)]%
        {rebuffi2017icarl}
\bibfield{author}{\bibinfo{person}{Sylvestre-Alvise Rebuffi}, \bibinfo{person}{Alexander Kolesnikov}, \bibinfo{person}{Georg Sperl}, {and} \bibinfo{person}{Christoph~H Lampert}.} \bibinfo{year}{2017}\natexlab{}.
\newblock \showarticletitle{icarl: Incremental classifier and representation learning}. In \bibinfo{booktitle}{\emph{Proceedings of the IEEE conference on Computer Vision and Pattern Recognition}}. \bibinfo{pages}{2001--2010}.
\newblock


\bibitem[Rossler et~al\mbox{.}(2019)]%
        {Rossler20191}
\bibfield{author}{\bibinfo{person}{Andreas Rossler}, \bibinfo{person}{Davide Cozzolino}, \bibinfo{person}{Luisa Verdoliva}, \bibinfo{person}{Christian Riess}, \bibinfo{person}{Justus Thies}, {and} \bibinfo{person}{Matthias Niessner}.} \bibinfo{year}{2019}\natexlab{}.
\newblock \showarticletitle{FaceForensics++: Learning to detect manipulated facial images}.
\newblock \bibinfo{journal}{\emph{Proceedings of the IEEE International Conference on Computer Vision}}  \bibinfo{volume}{2019-October} (\bibinfo{year}{2019}), \bibinfo{pages}{1 – 11}.
\newblock
\urldef\tempurl%
\url{https://doi.org/10.1109/ICCV.2019.00009}
\showDOI{\tempurl}


\bibitem[Semola et~al\mbox{.}(2022)]%
        {semola2022continual}
\bibfield{author}{\bibinfo{person}{Rudy Semola}, \bibinfo{person}{Vincenzo Lomonaco}, {and} \bibinfo{person}{Davide Bacciu}.} \bibinfo{year}{2022}\natexlab{}.
\newblock \showarticletitle{Continual-learning-as-a-service (claas): On-demand efficient adaptation of predictive models}.
\newblock \bibinfo{journal}{\emph{arXiv preprint arXiv:2206.06957}} (\bibinfo{year}{2022}).
\newblock


\bibitem[Verdoliva(2020)]%
        {Verdoliva2020910}
\bibfield{author}{\bibinfo{person}{Luisa Verdoliva}.} \bibinfo{year}{2020}\natexlab{}.
\newblock \showarticletitle{Media Forensics and DeepFakes: An Overview}.
\newblock \bibinfo{journal}{\emph{IEEE Journal on Selected Topics in Signal Processing}} \bibinfo{volume}{14}, \bibinfo{number}{5} (\bibinfo{year}{2020}), \bibinfo{pages}{910 – 932}.
\newblock
\urldef\tempurl%
\url{https://doi.org/10.1109/JSTSP.2020.3002101}
\showDOI{\tempurl}


\bibitem[Wang et~al\mbox{.}(2023)]%
        {wang2023open}
\bibfield{author}{\bibinfo{person}{Jun Wang}, \bibinfo{person}{Omran Alamayreh}, \bibinfo{person}{Benedetta Tondi}, {and} \bibinfo{person}{Mauro Barni}.} \bibinfo{year}{2023}\natexlab{}.
\newblock \showarticletitle{Open Set Classification of GAN-based Image Manipulations via a ViT-based Hybrid Architecture}. In \bibinfo{booktitle}{\emph{Proceedings of the IEEE/CVF Conference on Computer Vision and Pattern Recognition}}. \bibinfo{pages}{953--962}.
\newblock


\bibitem[Wang et~al\mbox{.}(2020)]%
        {wang2020cnn}
\bibfield{author}{\bibinfo{person}{Sheng-Yu Wang}, \bibinfo{person}{Oliver Wang}, \bibinfo{person}{Richard Zhang}, \bibinfo{person}{Andrew Owens}, {and} \bibinfo{person}{Alexei~A Efros}.} \bibinfo{year}{2020}\natexlab{}.
\newblock \showarticletitle{CNN-generated images are surprisingly easy to spot... for now}. In \bibinfo{booktitle}{\emph{Proceedings of the IEEE/CVF conference on computer vision and pattern recognition}}. \bibinfo{pages}{8695--8704}.
\newblock


\bibitem[Wu et~al\mbox{.}(2020)]%
        {wu2020watermarking}
\bibfield{author}{\bibinfo{person}{Hanzhou Wu}, \bibinfo{person}{Gen Liu}, \bibinfo{person}{Yuwei Yao}, {and} \bibinfo{person}{Xinpeng Zhang}.} \bibinfo{year}{2020}\natexlab{}.
\newblock \showarticletitle{Watermarking neural networks with watermarked images}.
\newblock \bibinfo{journal}{\emph{IEEE Transactions on Circuits and Systems for Video Technology}} \bibinfo{volume}{31}, \bibinfo{number}{7} (\bibinfo{year}{2020}), \bibinfo{pages}{2591--2601}.
\newblock


\bibitem[Zhang et~al\mbox{.}(2024)]%
        {zhang2024joint}
\bibfield{author}{\bibinfo{person}{Yibo Zhang}, \bibinfo{person}{Weiguo Lin}, {and} \bibinfo{person}{Junfeng Xu}.} \bibinfo{year}{2024}\natexlab{}.
\newblock \showarticletitle{Joint Audio-Visual Attention with Contrastive Learning for More General Deepfake Detection}.
\newblock \bibinfo{journal}{\emph{ACM Transactions on Multimedia Computing, Communications and Applications}} \bibinfo{volume}{20}, \bibinfo{number}{5} (\bibinfo{year}{2024}), \bibinfo{pages}{1--23}.
\newblock


\bibitem[Zhao et~al\mbox{.}(2023)]%
        {zhao2023proactive}
\bibfield{author}{\bibinfo{person}{Yuan Zhao}, \bibinfo{person}{Bo Liu}, \bibinfo{person}{Ming Ding}, \bibinfo{person}{Baoping Liu}, \bibinfo{person}{Tianqing Zhu}, {and} \bibinfo{person}{Xin Yu}.} \bibinfo{year}{2023}\natexlab{}.
\newblock \showarticletitle{Proactive deepfake defence via identity watermarking}. In \bibinfo{booktitle}{\emph{Proceedings of the IEEE/CVF winter conference on applications of computer vision}}. \bibinfo{pages}{4602--4611}.
\newblock


\bibitem[Zi et~al\mbox{.}(2020)]%
        {Zi20202382}
\bibfield{author}{\bibinfo{person}{Bojia Zi}, \bibinfo{person}{Minghao Chang}, \bibinfo{person}{Jingjing Chen}, \bibinfo{person}{Xingjun Ma}, {and} \bibinfo{person}{Yu-Gang Jiang}.} \bibinfo{year}{2020}\natexlab{}.
\newblock \showarticletitle{WildDeepfake: A Challenging Real-World Dataset for Deepfake Detection}.
\newblock \bibinfo{journal}{\emph{MM 2020 - Proceedings of the 28th ACM International Conference on Multimedia}} (\bibinfo{year}{2020}), \bibinfo{pages}{2382 – 2390}.
\newblock
\urldef\tempurl%
\url{https://doi.org/10.1145/3394171.3413769}
\showDOI{\tempurl}


\end{thebibliography}
\end{document}